\pdfoutput=1
\documentclass[11pt]{article}

\usepackage[preprint]{acl}

\usepackage{times}
\usepackage{latexsym}
\usepackage{amsmath}
\usepackage{subcaption}
\usepackage{hyperref}       % hyperlinks
\usepackage{url}            % simple URL typesetting
\usepackage{booktabs}       % professional-quality tables
\usepackage{amsfonts}       % blackboard math symbols
\usepackage{nicefrac}       % compact symbols for 1/2, etc.
\usepackage{xcolor}         % colors
\usepackage{mdframed}
\usepackage{threeparttable}
\usepackage{graphics}
\usepackage{multirow}
\usepackage{rotating}
\usepackage{array}
\usepackage{caption}
\usepackage{xcolor}
\usepackage{colortbl}
\usepackage{listings}
\usepackage{enumitem}
\usepackage{amsmath}
\usepackage{cleveref}
\usepackage{hyperref}
\usepackage{xspace}
\definecolor{grey}{rgb}{0.5, 0.5, 0.5}
\usepackage[T1]{fontenc}
\usepackage[utf8]{inputenc}

\usepackage{microtype}

\usepackage{inconsolata}

\usepackage{graphicx}

\title{Attention Score is not All You Need for Token Importance \\ Indicator  in  KV Cache Reduction: Value Also Matters }

\author{Zhiyu Guo, Hidetaka Kamigaito, Taro Watanabe \\
  Nara Institute of Science and Technology \\
  \texttt{\{guo.zhiyu.fy1, kamigaito.h, taro\}@is.naist.jp} \\}

\begin{document}
\maketitle
\begin{abstract}
Scaling the context size of large language models (LLMs) enables them to perform various new tasks, e.g., book summarization. However, the memory cost of the Key and Value (KV) cache in attention significantly limits the practical applications of LLMs. Recent works have explored token pruning for KV cache reduction in LLMs, relying solely on attention scores as a token importance indicator. However, our investigation into value vector norms revealed a notably non-uniform pattern questioning their reliance only on attention scores. Inspired by this, we propose a new method: Value-Aware Token Pruning (VATP) which uses both attention scores and the $ \ell_{1} $ norm of value vectors to evaluate token importance. Extensive experiments on LLaMA2-7B-chat and Vicuna-v1.5-7B across 16 LongBench tasks demonstrate that VATP outperforms attention-score-only baselines in over 12 tasks, confirming the effectiveness of incorporating value vector norms into token importance evaluation of LLMs.\footnote{Code is 
 available at: \href{https://github.com/guozhiyu/vatp}{https://github.com/guozhiyu/vatp}}
% \\vatp\_kvcache
\end{abstract}

\section{Introduction}

Recent studies have focused on scaling the context sizes of Transformer-based \cite{vaswani2017attention} large language models (LLMs) in addition to scaling data, compute, and model size. For example, the context size has increased from 2048 tokens in GPT-3 \cite{brown2020language} and LlaMA1 \cite{touvron2023llamaa} to 2 million tokens in Gemini 1.5 Pro \cite{reid2024gemini}. Longer context sizes enable LLMs to address tasks that extend beyond conventional capabilities, such as book-length summarization \cite{chang2024booookscore}, SWE-agent \cite{yang2024sweagent}, and many-shot in-context learning \cite{agarwal2024many}. However, the enormous inference costs of LLMs limit their applications. Therefore, in addition to model weight compression \cite{dettmers2022gptint,sun2024a}, enhancing the efficiency of long-context inference has become increasingly important.

LLMs utilize an auto-regressive framework in which tokens are produced sequentially. The generation of each token relies on the tokens generated before it. During generation, the key and value tensors of previously generated tokens, known as the KV cache, have to be preserved in memory throughout the generation process for attention computation. The memory cost of the KV cache scales linearly to the batch size and sequence length. This prohibitive memory cost has become a critical bottleneck limiting the applications of long-context LLMs. 

\begin{figure*}[ht]
  \centering
  \begin{subfigure}{0.32\linewidth}
    \includegraphics[width=\linewidth]{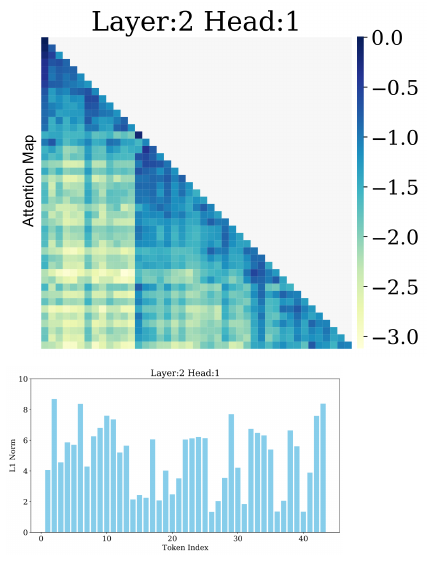}
    \caption{Layer 1-2}
    \label{fig:subfig1}
  \end{subfigure}\hfill
  \begin{subfigure}{0.3\linewidth}
    \includegraphics[width=\linewidth]{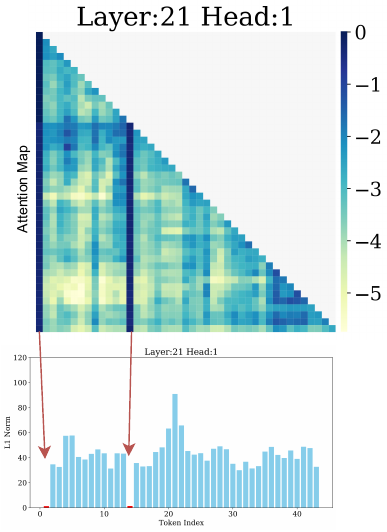}
    \caption{Layer 3-31}
    \label{fig:subfig2}
  \end{subfigure}\hfill
  \begin{subfigure}{0.3\linewidth}
    \includegraphics[width=\linewidth]{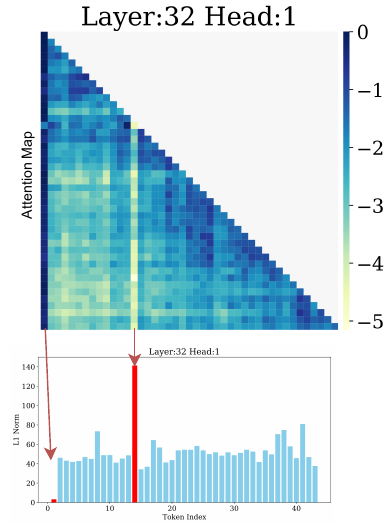}
    \caption{Layer 32}
    \label{fig:subfig3}
  \end{subfigure}
  \caption{Typical attention map (logarithmic) and value vector norm patterns in LLaMA2-7B-chat. Key observations include: (1) The \( \ell_{1} \) norms are non-uniformly distributed across tokens in all layers and heads. (2) In figure (b), for most heads in layers 3-31, regardless of the input text, there are two attention sink \cite{xiao2024efficient} tokens at the beginning of the text. Contrary to their massive attention scores, their \( \ell_{1} \) norms are close to 0 (highlighted in red). (3) In some heads of the last layer, the second attention sink token in figure (b) has a smaller attention score than other tokens, while its \( \ell_{1} \) norm is significantly larger than those of other tokens.}

  \label{fig:mainfigure}
\end{figure*}

One of the approaches for improving long-context inference efficiency is token pruning, which has been extensively explored for BERT \cite{goyal2020power,zhao-etal-2022-fine,guan-etal-2022-transkimmer}. However, these methods necessitate a complicated fine-tuning process to restore optimal performance. Given the extensive text corpora and the considerable size of LLMs, such fine-tuning becomes exceptionally challenging and less preferred. Fortunately, recent studies \cite{zhang2023ho,liu2023scissorhands,ge2024model,xiao2024efficient} have explored token pruning for KV cache reduction without the need for fine-tuning, indicating that a significant number of tokens can be pruned with minimal impact on performance during token generation. It is notable that these studies unanimously chose to rely solely on the attention score as the token importance indicator in LLMs. This choice is reasonable for LLMs, as training additional token importance predictor \cite{guan-etal-2022-transkimmer} is computationally expensive. 
% Nevertheless, before establishing the attention score as the default choice for the token importance indicator, we pose a timely question: \textit{Are there any essential elements that may have been accidentally omitted when considering pivotal tokens for KV cache reduction?}

In the pre-LLM era, however, the reliability of attention score as indicator of token importance was widely questioned \citep{wiegreffe-pinter-2019-attention, clark-etal-2019-bert, hassid2022much}.\footnote{For example, \citet{clark-etal-2019-bert} observed that special tokens tend to receive disproportionately large attention scores in BERT, yet those scores can often be significantly changed without impacting the model's predictions.} Despite these concerns, in the context of KV cache reduction for LLMs, attention score appears to be highly indicative of token importance. Recent studies have demonstrated that removing even a very small number of tokens with large attention scores can significantly degrade the model's performance \cite{zhang2023ho, xiao2024efficient}. Nevertheless, before establishing attention score as the default choice for the token importance indicator in LLMs, we pose a timely question: \textit{Are there any essential elements that may have been accidentally omitted when considering pivotal tokens for KV cache reduction?}

Since the output of the attention mechanism is the result of the multiplication of each token's attention score with its corresponding value vector, we investigated the value vectors of LLMs. We found the \( \ell_{1} \) norm of each token is highly non-uniformly distributed, showing distinct differences in magnitude.  Previous study \cite{xiao2024efficient} identifies the attention sink tokens with massive attention scores. We find, in contrast to the attention scores, the value vector norms of the attention sink tokens are much smaller than other tokens. Such a phenomenon is similar to the finding in small Transformer models \cite{kobayashi-etal-2020-attention}. When considering each token's effects on the attention output, their value vector should also be considered.

Building upon this observation, we introduce a new approach termed Value-Aware Token Pruning (VATP). Unlike traditional methods that rely solely on attention score, VATP augments the attention score with the norm of the value vector, providing a robust metric for evaluating token importance. Specifically, we propose a novel token pruning metric, where the KV cache of each token is assessed based on the product of its attention score and the \( \ell_{1} \) norm of the corresponding value vector. We conduct extensive experiments on the LLaMA2-7B-chat and Vicuna-v1.5-7B models, evaluating VATP across 16 long-context tasks from the LongBench \cite{bai2023longbench} benchmark. The results demonstrate that VATP outperforms attention-score-only baselines across a wide variety of tasks. \textit{Our research clearly reveals the critical, yet previously overlooked, role of the value vector norms in KV cache reduction, challenging the prevailing belief that attention score is all you need for evaluating token importance in LLMs.}

\begin{figure*}[ht]

\vskip 0.2in
\begin{center}

\includegraphics[width=0.9\textwidth]{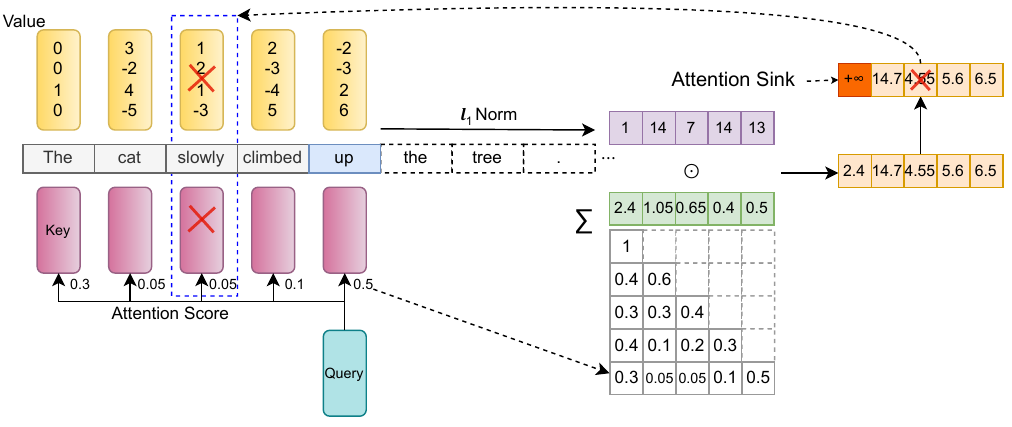}

     \centering
        \caption{ An illustration example of Value-Aware Token Pruning (VATP) method. VATP uses the product of value vector norm and attention score to evaluate the importance of each token's KV cache. The KV cache of the token with the lowest importance score is removed. Here we use the accumulated attention score proposed by H$_2$O \cite{zhang2023ho}, resulting in H$_2$O w/ VATP variant.}
        \label{methodill}
\end{center}   

\end{figure*}

\section{Related Work}
Many works have explored improving the inference efficiency of Transformer via token pruning. \citet{goyal2020power,zhao-etal-2022-fine} accelerate BERT by eliminating redundant word vectors based on attention scores.  \citet{zhao-etal-2022-fine} also reduces information loss in token pruning by using pooling strategies to compress the pruned tokens. To decide which tokens to prune, Transkimmer \cite{guan-etal-2022-transkimmer} adds an extra prediction module before each layer. Nevertheless, these methods are primarily designed for models similar to BERT \cite{devlin-etal-2019-bert}, which are encoder-only and typically smaller in size. They also generally require a complex fine-tuning process, making them less suitable for the currently prevalent larger auto-regressive language models.

Recent works investigate reducing the KV cache of unimportant tokens during auto-regressive generation. H$_2$O \cite{zhang2023ho} dynamically retain a combination of recent and heavy-hitter tokens in the KV cache, which are identified based on the accumulated attention scores. Scissorhands \cite{liu2023scissorhands} uses the attention scores from a history window as the token importance indicator. FastGen \cite{ge2024model} adjusts its compression strategies to align with the attention structure of each head. \citet{xiao2024efficient} observe attention sink tokens with massive attention scores, they simply keep the attention sink tokens together with the sliding window tokens to anchor the attention computation and stabilize the model's performance. Notably, these methods primarily focus on attention scores as the basis for determining which tokens to prune, our work for the first time explores the critical role of value vector norm in token pruning.

\section{Value-Aware Token Pruning}
This section introduces Value-Aware Token Pruning (VATP), starting from observations and concluding with our algorithm. A visual example of VATP is shown in Figure \ref{methodill}. 

\subsection{Observations}
The output of an attention head at step $t$ is defined as follows: 
\begin{equation}
\text{Attention}(Q, K, V)_t = \sum_{i \leq t} a_i^t \boldsymbol{v}_i
\label{atten}
\end{equation}
where $a_i^t$ is the attention score of query token $t$ to token $i$, and $\boldsymbol{v}_i$ is the value state of token $i$. The attention output for the token $t$ is thus a weighted sum of the value states $\boldsymbol{v}_i$ of all preceding tokens $i \leq t$, where the weights are the corresponding attention scores $a_i^t$.
The goal of token pruning is to remove tokens that have a minimal impact on the attention output. From Equation (\ref{atten}), each token's influence on the attention output is determined by both the attention score $a_i^t$ and the value vector $\boldsymbol{v}_i$. 

Here, we jointly analyze the attention maps and the corresponding value vectors. In Figure \ref{fig:mainfigure}, the $\ell_1$ norm of value vector exhibits a highly non-uniform distribution across all layers and heads. Notably, the two attention sink tokens\footnote{The first token is the starting word token, the second token is often the token representing the first period (.) or newline token (\textbackslash n) in the text, there are corresponding to massive activations as discussed in \citep{sun2024massive}. } often show a striking contrast between their attention scores and value vector norms. This observation is similar to the study of small Transformer models \citep{kobayashi-etal-2020-attention}.
% Most of the time, these two tokens have very massive attention scores, but value vector norms are close to 0. Occasionally, the opposite is true: the second token has very low attention score but high value vector norms. 

\subsection{Methodology}
The above observation highlights the importance of considering both the attention score and the value vector norm together. Such a dual consideration provides a more comprehensive understanding of each token's influence on the attention output. Consequently, it becomes obvious to implement
token pruning strategies that take into account attention score and value vector norm simultaneously. 
\paragraph{Attention Score} H$_2$O \cite{zhang2023ho} uses the accumulated attention score as token importance indicator. Specially, the token importance score for a given token $k$ at decoding step $t$ is calculated as:
\begin{equation}
S_k^t = \sum_{k \leq j \leq t} a_k^j
\label{h20}
\end{equation}
Scissorhands \cite{liu2023scissorhands} use the attention score based on the history window with size $w$. 
\begin{equation}
S_k^t = \sum_{\max(t-w, k) \leq j \leq t} a_k^j
\label{schand}
\end{equation}

\paragraph{Value-aware Pruning Metric} Motivated by the success of LLM weight pruning work Wanda \cite{sun2024a}, which evaluates model weight importance by the product of its magnitude and the corresponding input feature norm, we propose a new metric to evaluate token importance.  For each token in an attention head, its importance is evaluated by the product of its attention score $S_k^t$ and the corresponding value vector norm. Specifically, the score for the token $k$ at decoding step $t$ is defined by:
\begin{equation}
I_k^t = S_k^t  \cdot \left\| \boldsymbol{v}_k \right\|_1
\label{valeq}
\end{equation}
where $\left\| \boldsymbol{v}_k \right\|_1$ is the $\ell_1$ norm of of token $k$'s value vector. The attention score $S_k^t$ can be either Eq. (\ref{h20}) or Eq. (\ref{schand}). We empirically find $\ell_1$ norm performs better than $\ell_2$ norm in Appendix \ref{abla:norm}. The computation of VATP metric is straightforward by jointly considering the attention score and value vector.
\paragraph{Attention Sink Tokens} From the previous observations, the attention sink tokens have very small $\ell_1$ norm. Based on our metric, the importance scores of attention sink tokens are significantly downgraded, and they could be accidentally removed. While the value updates from those tokens may be small, the attention distribution of the rest tokens will be largely shifted after the removal, leading to deteriorated performance \cite{xiao2024efficient}. Thus we intentionally keep the first $\mathbf{F}$ tokens.

\begin{table*}[t]

\fontsize{22}{26}\selectfont
\setlength{\tabcolsep}{5pt}
\centering
\begin{threeparttable}

\scalebox{0.4}{
\begin{tabular}{l|lcccccccccccccccc}
\specialrule{1pt}{0pt}{2pt}
&\multirow{2}{*}{Method} & \multicolumn{3}{c}{Single-Document QA} & \multicolumn{3}{c}{Multi-Document QA}& \multicolumn{3}{c}{Summarization}& \multicolumn{3}{c}{Few-shot Learning}& \multicolumn{2}{c}{Synthetic} & \multicolumn{2}{c}{Code} \\
\cmidrule(lr){3-5}\cmidrule(lr){6-8}\cmidrule(lr){9-11}\cmidrule(lr){12-14}\cmidrule(lr){15-16}\cmidrule(lr){17-18}
&& 1-1 & 1-2 & 1-3 & 2-1 & 2-2 & 2-3 & 3-1 & 3-2 & 3-3 & 4-1 & 4-2 & 4-3 & 5-1 & 5-2 & 6-1 & 6-2 \\

\specialrule{1pt}{2pt}{2pt}
\multirow{6}{*}{\rotatebox[origin=c]{90}{\fontsize{18}{100}\selectfont LlaMA2-7B-chat}} & All Budget & 19.12 &20.99 & 37.55 &30.55 & 27.44 &8.31 &27.77 & 20.67 &24.39 & 58.33&86.22 & 39.14 &3.89 &9.67 & 59.88 & 48.61\\
% \cline{2-18}
 & StreamLLM & 15.4&18.6&25.96&28.19&23.59&7.08&23.87&19.97&22.52&56.67&86.45&38.72&3.87&2.62&58.55&48.28\\
 \cline{2-18}
 &H$_2$O & 18.4&18.83&33.67&\textbf{30.18} &25.74&7.85&\textbf{26.18}&21.12&23.44&\textbf{58.67}&85.35&\textbf{39.0}&4.37&7.0&59.4&49.09 \\
 % \rowcolor{gray!25}
 &\cellcolor{grey!10}w/ VATP & \cellcolor{grey!10}\textbf{18.77} & \cellcolor{grey!10}\textbf{19.6} & \cellcolor{grey!10}\textbf{35.31} & \cellcolor{grey!10}29.95 & \cellcolor{grey!10}\textbf{27.15} & \cellcolor{grey!10}\textbf{8.44} & \cellcolor{grey!10}26.08 & \cellcolor{grey!10}\textbf{21.14} & \cellcolor{grey!10}\textbf{23.76} & \cellcolor{grey!10}58.33 & \cellcolor{grey!10}\textbf{86.09} & \cellcolor{grey!10}38.74 & \cellcolor{grey!10}\textbf{4.39} & \cellcolor{grey!10}\textbf{8.33} & \cellcolor{grey!10}\textbf{59.56} & \cellcolor{grey!10}\textbf{49.46} \\
 \cline{2-18}
 & Scissorhands &18.5 &19.32&36.35&29.5&25.51&8.59&25.42&20.35&\textbf{23.86}&57.33&85.55&38.77&\textbf{4.38}&6.0&58.33&\textbf{48.86} \\
 &\cellcolor{grey!10}w/ VATP &\cellcolor{grey!10}\textbf{19.4} & \cellcolor{grey!10}\textbf{19.53} & \cellcolor{grey!10}\textbf{36.58} & \cellcolor{grey!10}\textbf{29.57} & \cellcolor{grey!10}\textbf{27.71} & \cellcolor{grey!10}\textbf{9.66} & \cellcolor{grey!10}\textbf{26.17} & \cellcolor{grey!10}\textbf{20.46} & \cellcolor{grey!10}23.63 & \cellcolor{grey!10}\textbf{58.0} & \cellcolor{grey!10}\textbf{85.98} & \cellcolor{grey!10}\textbf{38.9} & \cellcolor{grey!10}4.18 & \cellcolor{grey!10}\textbf{10.0} & \cellcolor{grey!10}\textbf{59.39} & \cellcolor{grey!10}48.71\\
\specialrule{1pt}{2pt}{10pt}\specialrule{1pt}{2pt}{2pt}

\multirow{6}{*}{\rotatebox[origin=c]{90}{\fontsize{18}{100}\selectfont vicuna-v1.5-7B}}&All Budget& 18.67&23.39&39.25&27.48&19.62&8.09&30.84&22.85&24.7&64.33&86.53&39.69&4.33&13.0&50.15&36.52\\
&StreamLLM&16.97&21.55&26.01&23.79&16.94&5.83&26.6&21.94&22.48&62.67&86.4&39.54&2.0&11.33&49.56&37.79\\
 \cline{2-18}
 &H$_2$O &18.46&21.84&32.99&26.86&\textbf{19.7}9&6.04&27.92&23.2&\textbf{23.78}&64.0&79.06&39.19&4.33&11.67&\textbf{51.38}&36.86\\
 &\cellcolor{grey!10}w/ VATP & \cellcolor{grey!10}\textbf{18.86} & \cellcolor{grey!10}\textbf{21.89} & \cellcolor{grey!10}\textbf{36.94} & \cellcolor{grey!10}\textbf{28.23} & \cellcolor{grey!10}19.47 & \cellcolor{grey!10}\textbf{7.72} & \cellcolor{grey!10}\textbf{28.57} & \cellcolor{grey!10}\textbf{23.21} & \cellcolor{grey!10}23.74 & \cellcolor{grey!10}\textbf{64.33} & \cellcolor{grey!10}\textbf{86.57} & \cellcolor{grey!10}\textbf{40.02} & \cellcolor{grey!10}4.33 & \cellcolor{grey!10}\textbf{13.0} & \cellcolor{grey!10}50.18 & \cellcolor{grey!10}\textbf{37.54}
\\
 \cline{2-18}
  & Scissorhands &18.19&20.56&33.83&26.08&18.93&5.99&26.14&22.62&23.24&61.67&80.26&39.7&4.33&9.0&\textbf{49.91}&35.05 \\
  &\cellcolor{grey!10}w/ VATP &\cellcolor{grey!10}\textbf{18.99} & \cellcolor{grey!10}\textbf{21.95} & \cellcolor{grey!10}\textbf{37.63} & \cellcolor{grey!10}\textbf{28.22} & \cellcolor{grey!10}\textbf{20.3} & \cellcolor{grey!10}\textbf{7.98} & \cellcolor{grey!10}\textbf{27.82} & \cellcolor{grey!10}\textbf{23.45} & \cellcolor{grey!10}\textbf{23.44} & \cellcolor{grey!10}\textbf{62.33} & \cellcolor{grey!10}\textbf{86.36} & \cellcolor{grey!10}\textbf{39.89} & \cellcolor{grey!10}4.33 & \cellcolor{grey!10}\textbf{13.0} & \cellcolor{grey!10}48.73 & \cellcolor{grey!10}\textbf{36.11} \\
   \cline{1-18}

\end{tabular}

}

\end{threeparttable}\vspace{0pt}
\caption{Performance of different token pruning methods on LongBench at 50\% KV cache budget. To streamline the text, following \cite{bai2023longbench}, we refer to the dataset as ID (eg., 1-1 map to NarrativeQA, 2-2 map to 2WikiMultihopQA); the mapping from ID to the dataset and evaluation metrics are available in Table \ref{datastas} of Appendix.}\label{tab:longbench}
\end{table*}

\section{Experiments}

\subsection{Settings}
\paragraph{Models}We use two open LLMs, LLaMA2-7B-chat \cite{touvron2023llamab} and Vicuna-v1.5-7B-16k \cite{zheng2023judging}. For LLaMA2-7B-chat, we set the max sequence length as 4K. For Vicuna-v1.5-7B-16k, we set the max sequence length as 8K due to GPU memory limitation. We conduct all experiments using one A6000 GPU.
\paragraph{Dataset} To extensively assess the effectiveness of our method in real-world scenarios, we select all the English tasks in LongBench \cite{bai2023longbench} as our evaluation benchmark. The Longbench benchmark consists of 16 English tasks, each containing between 150 and 500 samples. This benchmark encompasses a diverse array of long-text tasks, including question answering, text summarization, few-shot learning, synthetic tasks, and code completion. The detailed information about the dataset is in Table \ref{datastas} in Appendix \ref{datai}. We use the official task-specific prompts to evaluate task-wise performance of instruction-tuned LLMs.
\paragraph{Baselines}We choose several token pruning works: StreamLLM \cite{xiao2024efficient}, H$_2$O \cite{zhang2023ho}, Scissorhands \cite{liu2023scissorhands}. The full KV cache is used for assessing the performance degradation. VATP needs to use both attention score and value vector norm. The attention score calculation method can be derived from either H$_2$O or Scissorhands. Using the accumulated attention score results in the “H$_2$O w/ VATP”, while using the attention score based on the history window results in the “Scissorhands w/ VATP”. Implementation details are in Appendix \ref{implede}.

\subsection{Results}
\paragraph{Main results} While individual task results may exhibit variability, the aggregate results presented in Table \ref{tab:longbench}
 are more stable and reliable.
For the LlaMA2-7B-chat model, the VATP method surpasses H$_2$O in 12 out of 16 tasks and outperforms Scissorhands in 13 out of 16 tasks. Similarly, for the Vicuna-v1.5-7B-16k model, VATP exceeds H$_2$O in 12 out of 16 tasks and Scissorhands in 14 out of 16 tasks. Note that for the tasks where VATP does not surpass the baseline, its performance is still very comparable to the baseline. However, in certain tasks (e.g., 1-3 in Vicuna), VATP significantly outperforms the baseline. The overall performance gains demonstrate the effectiveness of our proposed method.

\begin{figure}[t]
    \centering
    \includegraphics[width=0.4\textwidth]{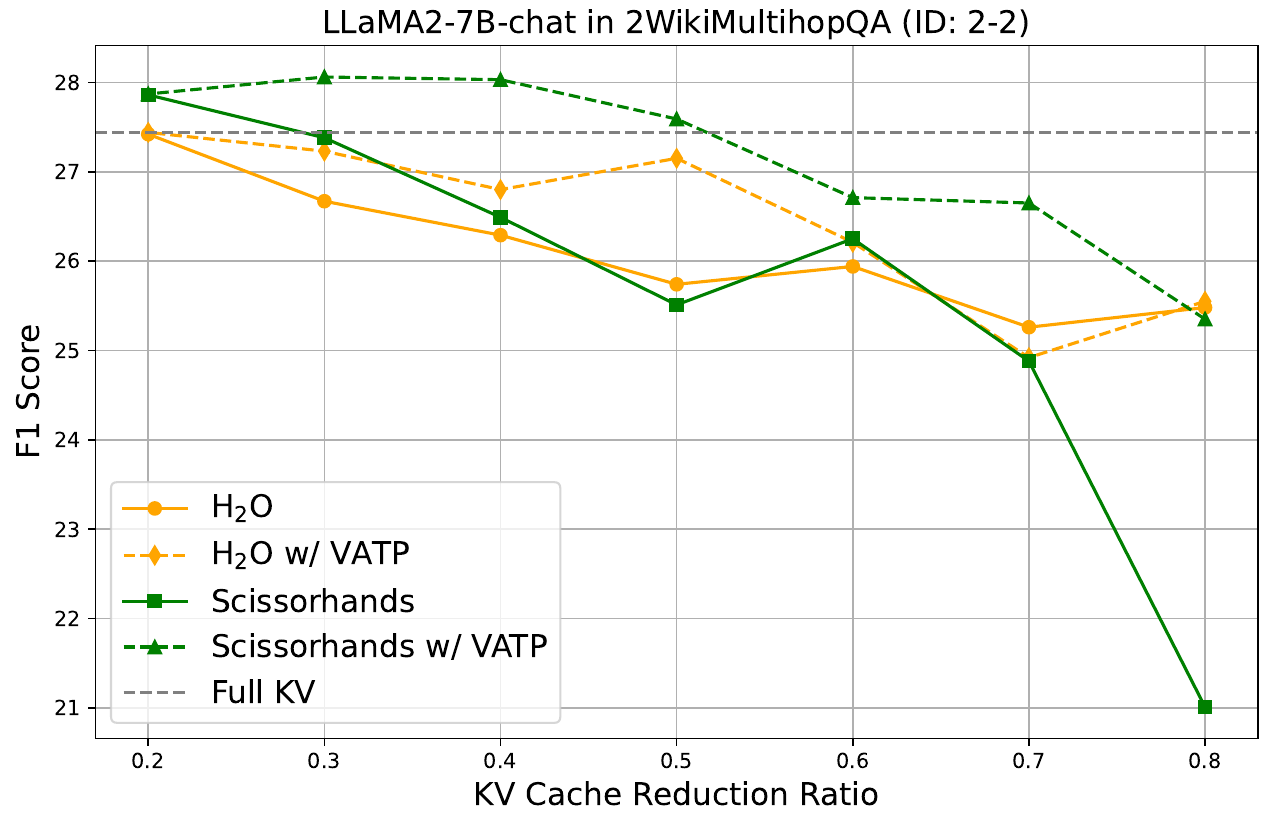}
    \caption{Performance on 2WikiMultihopQA of the LLaMA2-7B-chat with varying KV Cache Ratios.}
    \label{fig:performance_chart}
\end{figure}

\paragraph{KV budget ratio variation.} In Figure \ref{fig:performance_chart}, we test the performance on 2WikiMultihopQA at different KV cache reduction ratio. Scissorhands w. VATP yields the best performance in nearly all the KV budget ratios, in the high reduction ratio, it outperforms Scissorhands significantly. The improvement of H$_2$O w/VATP over H$_2$O is mainly at the less aggressive reduction ratio.

\vspace{-4pt}
\paragraph{Inference Efficiency.} VATP maintains the inherent simplicity of baseline methods by introducing negligible computation and memory overhead compared with H$_2$O and Scissorhands, since the size of the value vector norm is $\frac{1}{2d_{\text{head}}}$ of the KV cache size, ${d_{\text{head}}}=128$ for a 7B model. When integrating with H$_2$O, we need to calculate the accumulated attention score, this makes H$_2$O incompatible with FlashAttention~\citep{dao2022flashattention}. We implemented the integration of FlashAttention based on Scissorhands. As shown in Table \ref{inf_com} in Appendix \ref{inference}, VATP introduces no significant difference in generation throughput or memory usage compared to Scissorhands at the same KV budget. Detailed throughput improvements at different KV budgets are presented in Table \ref{throu_ipr}.

\section{Conclusion}
 This study addresses a critical yet previously overlooked aspect of token pruning in LLMs—the value vectors. Motivated by the observed highly non-uniform distribution of value vector norms, we propose a simple and easy-to-implement token pruning method called Value-Aware Token Pruning (VATP). VATP jointly considers both attention scores and value vector norms to assess token importance, introducing negligible computational overhead and requiring no additional fine-tuning. Extensive experiments demonstrate that VATP consistently outperforms attention-score-only approaches across a variety of long-context tasks. These findings provide fresh insights into the significant role of value vector norms in the context of KV cache reduction. This work paves the way for developing more advanced KV cache reduction algorithms, potentially leading to more efficient and scalable deployment of long-context LLMs in practical applications.

\section*{Limitations}
Our work has the following limitations:
\paragraph{FlashAttention Support for H$_2$O:} When integrating with H$_2$O, we need to calculate the accumulated attention score. However, the current implementation of FlashAttention~\citep{dao2022flashattention} does not return the attention matrix. Without integrating FlashAttention, the memory cost of prompt prefilling remains $O(n^2)$. Although it's unnecessary to store the attention matrix for all layers simultaneously, handling extensive context significantly increases the memory cost during prompt prefilling.  However, when integrating into Scissorhands, FlashAttention can be used. Since we don’t need to materialize the full attention matrix, Scissorhands only calculates the attention score based on last $w$ tokens. Scissorhands generally achieve better performance than H$_2$O in QA tasks. In addition, concurrent work \cite{devoto2024simple} shows a clear correlation between the $\ell_2$ norms of the key vector and the attention scores. This suggests the feasibility of reducing the KV cache without calculating attention scores, relying solely on the $\ell_2$ norms of key vectors. Pruning the KV cache jointly using the norms of both key and value vectors is an intriguing direction for future research.

\paragraph{Compatibility with grouped-query attention:} Similar to Scissorhands and H$_2$O, our method is currently not applicable to grouped-query attention (GQA) \cite{ainslie-etal-2023-gqa}. Token pruning and grouped-query attention are orthogonal in principle: grouped-query attention reduces the number of KV heads, while token pruning reduces the number of tokens. \citet{ren2024efficacy} use the group-wise averaged
attention score as the token importance score. Exploring the combination of VATP and GQA  represents a promising research direction.

\bibliography{custom}

\appendix
\section{Implementation Details}
For each input sequence, we set the KV cache budget to 50\% of the input prompt length. We assign uniform KV budgets across different heads and layers, as a uniform strategy is more practical to achieve actual inference improvements in hardware. FastGen \cite{ge2024model} uses a non-uniform strategy, thus we haven't chosen it as a baseline. For StreamLLM, we set the number of attention sink tokens to 20 for LLaMA2-7B-chat and 40 for Vicuna-v1.5-7B-16k. For VATP, we intentionally keep the first $\mathbf{F}$ tokens, where $\mathbf{F}=20$ for LLaMA2-7B-chat and $\mathbf{F}=40$ for Vicuna-v1.5-7B-16k. Given a KV budget of \( k \) tokens, the number of tokens selected by importance score in H$_2$O is \( \frac{k}{2} \), with a local window size also of \( \frac{k}{2} \). In Scissorhands, following \citet{liu2023scissorhands}, the number of tokens selected by importance score is \( k-10 \), with a local window size of 10 and a history window size of \( w=400 \). When integrating with Scissorhands and H$_2$O, the only differences are the token importance score and intentionally keeping attention sink tokens. 

\label{implede}

\section{Ablation Study}
\label{abla:norm}
Table \ref{table:norms} shows the F1 scores for 4 QA tasks under different types of norm for value vector: $\ell_1$, $\ell_2$, and $\ell_\infty$. Overall, the $\ell_1$ norm achieves the highest average performance with an average F1 score of 28.00, indicating that $\ell_1$ norm performs better across the evaluated tasks compared to $\ell_2$ and $\ell_\infty$ norms. Thus we use $\ell_1$ norm in all the experiments.
\begin{table}[h!]
\centering
\resizebox{\columnwidth}{!}{
\begin{tabular}{lccc}
\hline
Task & $\ell_1$ Norm &  $\ell_2$ Norm & $\ell_\infty$ Norm \\
\hline
Qasper & \textbf{19.60} & 18.47 & 18.76 \\
MultifieldQA (en) & 35.31 & \textbf{35.48} & 35.11 \\
HotpotQA & 29.95 & \textbf{30.30} & 30.20 \\
2WikiMQA & \textbf{27.15} & 26.55 & 26.76 \\
\hline
\textbf{Average} & \textbf{28.00} & 27.70 & 27.71 \\
\hline
\end{tabular}}
\caption{F1 Scores under Different Norms for 4 QA Tasks}
\label{table:norms}
\end{table}

\section{Inference Efficiency}
\label{inference}
\paragraph{Computation Overhead of VATP}
For long-context sequences, the computation and memory overhead of the full attention matrix becomes a significant challenge, making the integration of FlashAttention essential. Since H$_2$O requires materializing the entire attention matrix to compute the accumulated attention score, it is incompatible with FlashAttention. In contrast, Scissorhands only requires the attention scores of the last $w$ tokens, allowing us to avoid materializing the full attention matrix. We integrated FlashAttention with Scissorhands to maximize the efficiency gains of VATP.

To evaluate the computational overhead introduced by VATP, we measured the generation throughput on an A6000 Ada GPU with an input sequence length of 4096, an output length of 128, and a batch size of 12. The KV cache budget was set at 50\%. As shown in Table \ref{inf_com}, VATP introduces negligible computation and memory overhead compared to Scissorhands alone, with no significant difference in generation throughput or peak memory usage.
\begin{table}[h!]
\centering
\resizebox{\columnwidth}{!}{
\begin{tabular}{lcc}
\toprule
\textbf{} & \textbf{throughput (token/s)} & \textbf{Peak Memory} \\ 
\midrule
Scissorhand & 104.28 & 48380 MiB \\ 
w. VATP & 103.93 & 48380 MiB \\ 
\bottomrule
\end{tabular}}
\caption{Comparison of throughput and peak memory}
\label{inf_com}
\end{table}

\paragraph{Throughput improvement of VATP }
We evaluate the generation throughput (measured in tokens per second, tok/s) of Scissorhands with VATP across various KV cache budgets to analyze the performance benefits of our approach. The experiments are conducted on both A100-80G and A6000 Ada (49GB) GPUs to account for different hardware configurations. For these tests, we use an input sequence length of 4096 tokens and an output sequence length of 128 tokens. Here we use LLaMA2-7B-chat model.

As shown in Table \ref{throu_ipr}, the improvements in throughput across different hardware configurations underscore the practicality of VATP, particularly for real-world applications where long input sequences and memory constraints can be major bottlenecks. By integrating VATP, we can achieve more efficient token generation without sacrificing model accuracy, even when operating under constrained KV cache budgets.

\begin{table}[t!]
\centering
\begin{small}

\begin{tabular}{lccc}
\toprule
GPU & KV budget & Throughput (tok/s) & Speedup \\
\midrule
\multirow{3}{*}{A100} & 100\%  & 79.78 & 1.0× \\
                           & 50\% & 126.45 & 1.58× \\
                           & 25\%   & 179.97 & 2.26× \\ \hline
\multirow{3}{*}{A6000} & 100\%   & 68.02 & 1.0× \\
                       &50\%   & 103.93  & 1.53× \\
                       & 25\%   & 141.74  & 2.08× \\ 

\bottomrule
\end{tabular}
\caption{Generation throughput improvement of VATP at different KV budgets.}
\label{throu_ipr}
\end{small}
\end{table}

\section{Dataset Details}
We select the English subset from Longbench \cite{bai2023longbench}. Table \ref{datastas} shows the information of 16 tasks we use in the experiments.
\label{datai}

\begin{table*}[ht!]
\centering
\begin{tabular}{llrlr}
\toprule
ID & Dataset & Avg len & Metric & \#data \\
\midrule
1-1 & NarrativeQA & 18,409 & F1 & 200 \\
1-2 & Qasper & 3,619 & F1 & 200 \\
1-3 & MultiFieldQA-en & 4,559 & F1 & 150 \\
\midrule
2-1 & HotpotQA & 9,151 & F1 & 200 \\
2-2 & 2WikiMultihopQA & 4,887 & F1 & 200 \\
2-3 & MuSiQue & 11,214 & F1 & 200 \\
\midrule
3-1 & GovReport & 8,734 & Rouge-L & 200 \\
3-2 & QMSum & 10,614 & Rouge-L & 200 \\
3-3 & MultiNews & 2,113 & Rouge-L & 200 \\
\midrule
4-1 & TREC & 5,177 & Accuracy (CLS) & 200 \\
4-2 & TriviaQA & 8,209 & F1 & 200 \\
4-3 & SAMSum & 6,258 & Rouge-L & 200 \\
\midrule
5-1 & PassageCount & 11,141 & Accuracy (EM) & 200 \\
5-2 & PassageRetrieval-en & 9,289 & Accuracy (EM) & 200 \\
\midrule
6-1 & LCC & 1,235 & Edit Sim & 500 \\
6-2 & RepoBench-P & 4,206 & Edit Sim & 500 \\
\bottomrule
\end{tabular}
\caption{The dataset statistics in LongBench include several key metrics. 'Source' indicates where the context originates. The 'Avg len' (average length) is measured by the number of words for datasets in English (or code). 'Accuracy (CLS)' represents classification accuracy, while 'Accuracy (EM)' denotes exact match accuracy.}
\label{datastas}
\end{table*}

\end{document}